\documentclass[sigconf, edbt]{acmart-edbt2023}
\usepackage{algorithm}
\usepackage{caption}
\usepackage{subcaption}
\usepackage{float}
\usepackage{tablefootnote}

\usepackage{bbm}
\usepackage{balance}

\newcommand{\revise}[1]{\textcolor{black}{\textsc{} #1}}

\def\BibTeX{{\rm B\kern-.05em{\sc i\kern-.025em b}\kern-.08em
    T\kern-.1667em\lower.7ex\hbox{E}\kern-.125emX}}

\setcopyright{rightsretained}

\acmDOI{}

\acmISBN{978-3-89318-088-2}

\acmConference[EDBT 2023]{26th International Conference on Extending Database Technology (EDBT)}{28th March-31st March, 2023}{Ioannina, Greece}
\acmYear{2023}

\begin{document}

\title{Unsupervised Space Partitioning for Nearest Neighbor Search}

\author{Abrar Fahim}
\affiliation{%
  \institution{Bangladesh University of Engineering and Technology}
  \city{Dhaka}
  \country{Bangladesh}}
\email{1605075@ugrad.cse.buet.ac.bd}

\author{Mohammed Eunus Ali}
\affiliation{%
  \institution{Bangladesh University of Engineering and Technology}
 \city{Dhaka}
  \country{Bangladesh}}
\email{eunus@cse.buet.ac.bd}

\author{Muhammad Aamir Cheema}
\affiliation{%
  \institution{Faculty of Information Technology, Monash University}
  \country{Australia}}
\email{aamir.cheema@monash.edu}

\begin{abstract}
  Approximate Nearest Neighbor Search (ANNS) in high dimensional spaces is crucial for many real-life applications (e.g., e-commerce, web, multimedia, etc.) dealing with an abundance of data. This paper proposes an end-to-end learning framework that couples the partitioning (one critical step of ANNS) and learning-to-search steps using a custom loss function. A key advantage of our proposed solution is that it does not require any expensive pre-processing of the dataset, which is one of the critical limitations of the state-of-the-art approach. We achieve the above edge by formulating a multi-objective custom loss function that does not need ground truth labels to quantify the quality of a given data-space partition, making it entirely unsupervised. We also propose an ensembling technique by adding varying input weights to the loss function to train an ensemble of models to enhance the search quality. On several standard benchmarks for ANNS, we show that our method beats the state-of-the-art space partitioning method and the ubiquitous K-means clustering method while using fewer parameters and shorter offline training times. We also show that incorporating our space-partitioning strategy into state-of-the-art ANNS techniques such as ScaNN can improve their performance significantly. Finally, we present our unsupervised partitioning approach as a promising alternative to many widely used clustering methods, such as K-means clustering and DBSCAN.

\end{abstract}

\maketitle
\section{Introduction}
$K$-Nearest Neighbor Search ($k$-NNS) that finds the $k$ closest (or most similar) data points for a given query point in a high-dimensional space is a well-studied problem ~\cite{shakhnarovich_nearest-neighbor_2005, wang_learning_2015, andoni_approximate_2018, wang_hashing_2014}. The vast amount of high-dimensional data that applications have to deal with today and an ever-greater need to quickly search for relevant content necessitate a scalable and efficient search solution for many domains, including multimedia, e-commerce, and recommendation systems. Exact solutions to the $k$-NNS problem, where we seek the exact $k$ nearest neighbors, are challenging and computationally intractable due to the phenomenon of the \emph{curse of dimensionality} ~\cite{indyk_approximate_1998}. Thus, they are not practical for many applications. Recent research has shifted to Approximate Nearest Neighbors Search (ANNS) ~\cite{muja_fast_2009, andoni_approximate_2018, aumuller_ann-benchmarks_2018} to scale the NNS solution to larger datasets with more dimensions. ANNS aims to quickly find as many of the true nearest neighbors of the query point as possible by slightly trading off the returned answer's accuracy. This paper proposes an end-to-end unsupervised learning solution using neural networks to solve the ANNS problem.

The established way to search for the k-Nearest-Neighbors (k-NNs) is to first reduce the search space for finding the most relevant points using \emph{indexing} methods (such as KD-trees ~\cite{cayton_learning_2007}, quantization using K-means ~\cite{jegou_product_2011}, PCA trees ~\cite{abdullah_spectral_2014, sproull_refinements_1991}, LSH ~\cite{lv_multi-probe_2007, andoni_practical_2015} etc.), and then to speed up the search within those relevant points using \emph{sketching} methods (e.g., ScaNN ~\cite{guo_accelerating_2020}, ITQ ~\cite{gong_iterative_2013}, etc.). This paper focuses on improving the indexing part to speed up ANNS. Most existing indexing approaches rely on algorithmic constructions that are either entirely independent or only weakly dependent on the data distribution (e.g., KD-trees ~\cite{cayton_learning_2007}, LSH ~\cite{lv_multi-probe_2007, andoni_practical_2015}, random trees ~\cite{keivani_improved_2018, dasgupta_randomized_2013}). These approaches cannot correctly curate the created partitions to specific data distributions. Notably, K-means clustering, a simple and prominent approach for clustering used in the implementation of the state-of-the-art ANNS technique ScaNN ~\cite{guo_accelerating_2020}, can only form convex (mostly spherical) clusters of the dataset. These simple cluster shapes may not be sophisticated enough to represent more complex data distributions. 

Recently, there has been an increased interest in machine-learning-based solutions (particularly supervised learning) for index creation on the data to facilitate efficient search. Notably, ~\cite{kang_case_2021, kraska_case_2018} argue the case for \emph{learning} index structures and show the benefits and potential of replacing core components of database systems with learned models. 
A recent approach, \emph{Neural LSH} ~\cite{dong_learning_2020}, uses neural nets and graph partitioning to create a \emph{space partitioning} index, which divides the ambient space of the dataset into smaller parts. Neural LSH outperforms previous data partitioning baselines. Neural LSH first creates a $k$-NN graph from the dataset and then partitions the graph to divide the dataset into several bins using a combinatorial graph-partitioning algorithm ~\cite{sanders_think_2012}. Using the resulting graph partition, it trains a neural network to learn to classify new query points into specific bins of the partition. By assigning query points to specific bins, Neural LSH restricts the further search to the data points within the query's assigned bins to find the nearest neighbors. This approach has several shortcomings: (i) Ground truth labels needed to train the model are generated in a separate pre-processing step, (ii) the graph-partitioning algorithm used to create the ground truth labels takes hours on million-sized datasets, and most importantly, (iii) the neural network is only used to learn to classify query points into bins, with the partitioning step not forming a part of the learning pipeline. As a result, Neural LSH does not capitalize on the power of function approximation in creating space-partitioning indexes.

To address the limitations of traditional  (e.g., LSH, K-means clustering, etc.) and learning-based (e.g., Neural LSH) partitioning solutions, we propose an end-to-end learning solution for scalable and efficient ANNS. The key intuition of our approach is that we can create superior partitions of the dataset by having the neural network itself learns the partition in an unsupervised manner. We do this by devising a customized cost function, enabling the neural network to learn the partition without generating prior training labels. We also propose an ensemble approach that allows us to merge multiple complementary partitions to improve indexing performance. Even though we primarily design our approach to solve the ANNS problem,  without loss of generality, our unsupervised partitioning approach is a promising alternative to many widely used clustering methods like K-means clustering, DBSCAN ~\cite{ester_density-based_1996}, and spectral clustering ~\cite{ng_spectral_2001}. 

We conduct extensive experiments with two standard Nearest Neighbor Search (NNS) benchmark datasets ~\cite{aumuller_ann-benchmarks_2018}, which show that our proposed approach yields $5-10\%$ performance improvement over the current state-of-the-art models.  \revise{Moreover, we show that by incorporating our unsupervised space partitioning strategy, we can improve the performance of the current best-performing ANNS method, namely ScaNN, by approximately 40\%.}

In summary, our contributions in this work are as follows.
\begin{itemize}
    \item We introduce an end-to-end learning framework for learning partitions of the dataset without any expensive pre-processing steps.
    \item We couple the partitioning and learning stages into a single step to make both the components aware of each other, increasing the overall framework's training efficiency.
   
    \item We introduce a custom loss function that can score output partitions and is differentiable. This loss function is \emph{model-agnostic} and thus can be applied to any machine learning architecture (including neural networks) to learn a richer class of division boundaries.
    
    In our experiments (Section ~\ref{experiments}), we show that our loss function makes any model learn better partitions than those created by the baseline methods in most real-world settings. 
    \item We propose an ensembling technique by adding varying input weights to the loss function to train an ensemble of models to create \emph{multiple high-quality complementary partitions} of the same dataset, which enhances indexing performance. 
\end{itemize}

We organized the rest of this paper as follows: We first discuss some related work in the field of similarity search in Section ~\ref{related_work}. We then formally define the approximate $k$-nearest neighbor search problem in Section ~\ref{problem}. Then, in Section ~\ref{our_method}, we discuss our learning-based approach for space partitioning and Nearest Neighbor Search (NNS) in detail. We present our experiments by comparing the performance of our method with other space-partitioning baselines in Section ~\ref{experiments}. Finally, we close with a summary of our contributions in Section ~\ref{conclusion}.

\section{Related Work}\label{related_work}

The two major paradigms to solve the ANNS (or NNS) problem are \emph{indexing} and \emph{sketching}.

Indexing methods generally construct a data structure that, given a query point $q$, returns a subset of the dataset called a \emph{candidate set} that includes the nearest neighbors of the $q$. On the other hand, sketching methods compress the data points to compute approximate distances quickly~\cite{wang_learning_2015, wang_hashing_2014,sablayrolles_spreading_2019, liong_deep_2015}. The two paradigms are often combined in real-world applications to maximize the overall performance ~\cite{guo_accelerating_2020, wu_multiscale_2017, johnson_billion-scale_2017}.

\subsection{Sketching: Making Distance Computations Faster}

In the sketching approach, we compute a compressed representation of the data points to transform the dataset from $\mathbb{R}^d$ to $\mathbb{R}^{d^\prime}$, such that distances in $\mathbb{R}^d$ are preserved in $\mathbb{R}^{d^\prime}$. This transformation makes each distance computation between the query point and a data point easier since distances are now computed in $\mathbb{R}^{d^\prime}$ instead of in $\mathbb{R}^d$ ($d^\prime < d$). In order to find the nearest neighbors under this paradigm, the whole dataset (compressed version) still needs to be scanned and distances computed between all points in the dataset and the query point.  

Machine learning methods have been instrumental in the sketching approach. Most machine learning methods use a fairly simple optimization objective to minimize reconstruction error in the lower dimensional space to preserve distances in the higher dimensional space. There have been many such works under "Learning to Hash." ~\cite{wang_learning_2015, wang_hashing_2014}. We highlight the recent work ~\emph{ScaNN} ~\cite{guo_accelerating_2020}, which develops a novel quantization loss function that outperforms previous sketching methods and forms the current state-of-the-art in the sketching domain.

\subsection{Indexing: Reducing the Search Space}

Under the indexing paradigm, we discuss graph-based and space-partitioning approaches. We then explore the benefits of  \emph{learning} space-partitions for indexing.

\subsubsection{Graph-Based Approaches}
Graph-based algorithms are one class of algorithms that reduce the number of points to search through. Graph-based algorithms ~\cite{fu_fast_2018, hajebi_fast_2011, harwood_fanng_2016, malkov_efficient_2020} construct a graph from the dataset (can be a $k$-NN graph) and then perform a greedy walk for each query, eventually converging on the nearest neighbor(s). While graph-based methods are very fast, they have suboptimal locality of reference and access the datasets adaptively in rounds. This makes graph search not ideal in modern distributed systems that often store the data points in an external storage medium since access to that medium could be very slow relative to searching and processing indices of data points ~\cite{dong_learning_2020}.

\subsubsection{Space Partitioning Methods}
Another class of algorithms is space-partitioning algorithms. These methods partition the search space into several \emph{bins} by dividing the ambient space of the dataset $\mathbb{R}^{d}$.
In this paper, we focus on the space-partitioning approach. Given a query point $q$, we identify the bin containing $q$ and produce a list of nearby candidates from the data points present in the same bin (or, to boost the k-NN recall, in nearby bins as well).

Space partitioning methods have numerous benefits ~\cite{dong_learning_2020}. First, they are naturally applicable in distributed settings, where different machines can store points in different bins. Furthermore, each machine can do a nearest neighbor search locally using other NNS methods to speed up the search further. Finally, unlike graph-based methods, space partitioning/data clustering methods only access the data points in one shot, only requiring access to the dataset points once it finds a candidate set and identifies the relevant points within it. 

Popular space partitioning methods include LSH ~\cite{lv_multi-probe_2007, andoni_practical_2015, dasgupta_neural_2017}, Quantization-based approaches, where partitions are obtained using K-Means clustering of the dataset ~\cite{jegou_product_2011}, and tree-based approaches such as random-projection or PCA trees ~\cite{sproull_refinements_1991, bawa_lsh_2005, dasgupta_randomized_2013, keivani_improved_2018}.

Classical space-partitioning algorithms like LSH ~\cite{lv_multi-probe_2007, andoni_practical_2015, dasgupta_neural_2017}, KD-trees, and random projection trees ~\cite{dasgupta_random_2008, dasgupta_randomized_2013} cannot effectively optimize a partition to a specific data distribution. In our experiments in Section ~\ref{experiments}, we show that these approaches (especially LSH and random trees projection trees) perform poorly compared to the other baselines. To create partitions better tailored to individual data distributions, we now look into \emph{learning} based methods for space partitioning.

\subsection{Learning Indexes for Space Partitioning}

There has been some prior work on incorporating machine learning techniques to improve space partitioning in ~\cite{cayton_learning_2007, ram_which_2013, li_learning_2011}. We highlight in particular the work in ~\cite{li_learning_2011}, termed ~\emph{Boosted Search Forest}, which introduces a custom loss function similar to our method. However, Boosted Search Forest, like ~\cite{cayton_learning_2007} and ~\cite{ram_which_2013}, can only learn ~\emph{hyperplane} partitions to split the dataset. This limits their partitioning performance as hyperplanes may not be sufficient to split more sophisticated data distributions. In contrast, our loss allows any machine learning model to learn a wider class of partitions for a dataset. Moreover, using our loss, even a simple logistic regression model can learn better hyperplane partitions than these prior learning approaches, indicating that our loss function can better score partitions than the loss used in Boosted Search Forest.

A recent relevant work, ~\emph{Neural LSH} ~\cite{dong_learning_2020} uses supervised learning with neural networks to create a space partitioning index by first creating a k-NN graph of the input dataset and running a combinatorial graph partitioning algorithm to obtain a balanced graph partition. The graph partition divides the dataset into several bins. It then trains the neural network to correctly classify out-of-sample query points to specific bins of the partition.

Apart from the above, other notable recent works on learned indexes such as ~\emph{Flood} ~\cite{nathan_learning_2020} and ~\emph{Tsunami} ~\cite{nathan_learning_2020} are summarized in ~\cite{al-mamun_tutorial_2020}. 
While these learned indexes are very efficient, they do not scale well to high-dimensional datasets, which is our focus in this paper.

\section{Problem Definition}

\label{problem}

Let $\mathbb{R}^{d} $ be a $d$-dimensional space. Given a dataset $X = \{p_1, ..., p_n\}$ of size $n$ in $\mathbb{R}^{d} $ and a query point $q \in \mathbb{R}^{d}$, $k$-\emph{nearest neighbor search} returns the top-$k$ ranked points from $X$ that are the most \emph{similar} to the query point $q$. We can use the Euclidean distance or any custom distance function to define the distance between any two points, $x$ and $y$, in the data space. For example, if the distance function $D$ is Euclidean distance, then we define the distance between $q$ and data point $p_i$ as $D(q, p_i) = \sqrt{ (q^1 - p_i^1)^2 + (q^2 - p_i^2)^2 +\cdots+ (q^d - p_i^d)^2}$. In modern large-scale applications, either $n$, $d$, or both are large, with $n$ often being billions or more. \revise{When answering nearest neighbor queries in real-time,} explicitly computing $D(q, p_i)$ for all points in the dataset can be prohibitively expensive. If $n$ is large, traversing the whole dataset to find $k$-NN is intractable, and if $d$ is large, computing the $D$ function itself is time-consuming for each data point.

Thus, in \emph{Approximate $k$-Nearest Neighbor Search (ANNS)}, we relax the requirement of retrieving the \emph{exact} top-k ranked points from $X$ w.r.t $q$. In ANNS, we return $k$ points close to $q$, ensuring that as many of them are the true $k$-nearest neighbors of $q$ as possible. Let $N^{\prime}_k(q)$ be the answer set of $k$ data points returned by the ANNS, and $N_k(q)$ be the answer set of true $k$-NN for the query point $q$. Thus, in the ANNS, we aim to maximize \revise{\emph{k-NN accuracy}} of the answer set, where, 
\begin{equation}
    \revise{\text{k-NN accuracy} = \frac{|N^{\prime}_k(q) \bigcap N_k(q)|}{k}}
\end{equation}

\section{Our Method}\label{our_method}

This section presents the details of our proposed method to solve the ANNS problem using an unsupervised learning-based approach. First, we give a high-level overview of the proposed approach. We then discuss the details of the different core components of the system. Finally, we present a couple of enhancements that include ensembling and hierarchical partitioning schemes.

In this work, we improve upon the state-of-the-art partitioning method \emph{Neural LSH} ~\cite{dong_learning_2020}. Neural LSH takes hours to preprocess a million-sized dataset to generate training labels to pass to the neural network. In contrast, our model takes less than two hours to learn high-quality partitions, even on constrained hardware resources. More importantly, Neural LSH does not use the neural network to create the partitions themselves. We introduce an end-to-end learning method that uses a novel loss function to create dataset partitions and learn to classify out-of-sample queries in a single learning step. 

\subsection{Overview}
\begin{figure*}
    \centering
    \includegraphics[width=\textwidth]{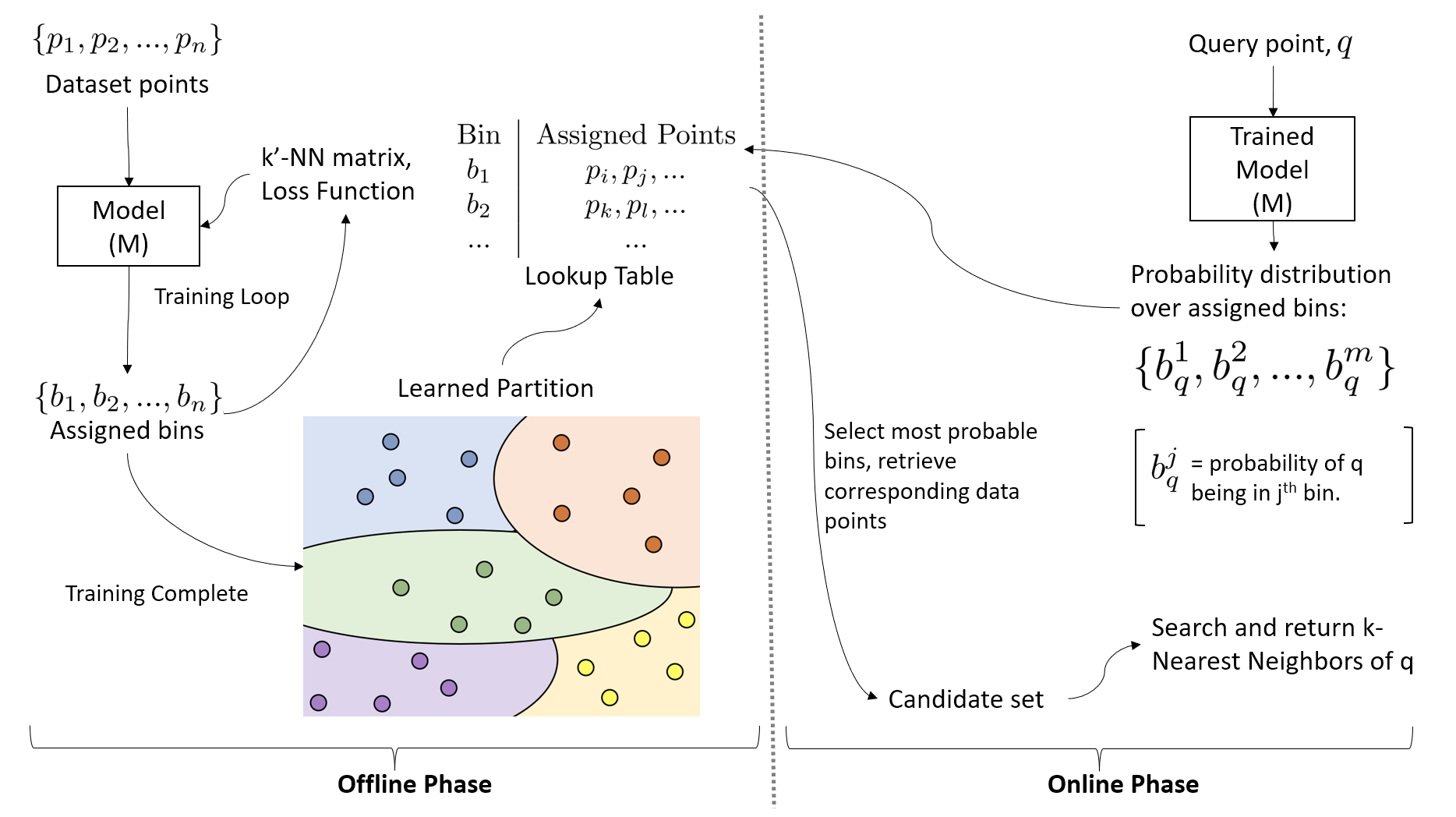}
    \caption{Overview of our method.}
    \label{overview}
\end{figure*}

We present a high-level overview of our proposed approach in Figure ~\ref{overview}. In general, the ANNS consists of two distinct phases, (i) the offline phase, where we train the model to partition the dataset, and (ii) the online phase, to answer queries in real-time using the trained model. 

In the offline phase, we use the dataset points in $X$, the NN matrix (described in Section~\ref{preprocessing}), and the loss function (described in Section~\ref{loss}) to train the model in the training loop. The trained model is then used to partition the dataset and create a lookup table to speed up the retrieval of candidate sets in the online phase. In the online phase, the trained model identifies the most likely bins to which the query $q$ belongs. The dataset points inside these bins are retrieved using the lookup table to form a candidate set of points containing probable nearby points of $q$. Finally, we search the reduced point sets in the candidate set to find the ANN of $q$. 

\subsection{The Offline Phase}
\label{offline} 

In the offline phase, we use the $n$ points in the dataset $X$ to train a machine learning model, $M$, to create a partition of the data space into $m$ bins.
\subsubsection{Preprocessing}
\label{preprocessing}

In this step, we create a $k^\prime$-NN matrix~\footnote{Note that this $k^\prime$ can be different from the $k$ used at query time for finding the approximate $k$ nearest neighbors w.r.t. the query.} from the dataset $X$. \revise{The $i^{th}$ row of the $k^\prime$-NN matrix contains the $k^\prime$ nearest neighbors of $p_i$ from $X$.}

This matrix captures the geometry and distribution of $X$ and provides this information to the model and the loss function. The $k^\prime$-NN matrix is essentially a $k^\prime$-NN graph many indexing methods use, represented as an adjacency list. Figure ~\ref{knn_mat} shows the representation of the $k^\prime$-NN matrix, where $p_i$ represents the $i^{th}$ point in $X$. The $i^{th}$ row in the matrix corresponds to all the  $k^\prime$ NNs of the $p_i$. The matrix shown is a 5-NN matrix: Each row contains the five nearest neighbors of the corresponding point. Note that this is the only preprocessing in our proposed approach. 

\begin{figure}
    \centering
    
    \begin{tabular}{ c|c c c c c} 
    
         $p_0$ & $p_7$ & $p_{10}$ & $p_3$ & $p_{21}$ & $p_{11}$\\ 
         $p_1$ & $p_0$ & $p_{20}$ & $p_{19}$ & $p_7$ & $p_5$ \\ 
         $p_2$ & $p_4$ & $p_9$ & $p_{20}$ & $p_{17}$ & $p_8$\\ 
         ... & ... & ... & ... & ... & ... \\

\end{tabular}
    \caption{5-NN matrix created from the dataset before model training in the offline phase.}
    \label{knn_mat}
    
\end{figure}
\revise{Preparing this matrix takes approximately 30 minutes on the million-sized dataset we used in our experiments. We compute all pairwise distances by traversing the whole dataset only once in the offline phase. In practical applications, the $k^\prime$-NN matrix is computed in the offline phase beforehand and stored in disk/cache for fast retrieval.}

\subsubsection{The Loss Function}
~\label{loss}
In this section, we discuss our proposed loss function, which is the key to our \emph{unsupervised} learning-based solution. The key intuition of the custom loss function to obtain a quality dataset partition comes from the following two objectives:
\begin{enumerate}
    \item \emph{Quality of candidate sets generated}: Intuitively, for a given query point $q$, a high-quality candidate set would have most or all of the nearest neighbors of $q$ contained within the candidate set.
    \item \emph{Even distribution of data points among all bins}: Ensuring even distribution of the $n$ data points among all the $m$ bins of the partition (roughly $n/m$ points per bin) results in smaller candidate set sizes generated per query on average. We desire fewer points per candidate set $(C)$ since the candidate set size $|C|$ is proportional to computation cost: We need to iterate through the points in $C$ to return the nearest neighbors of $q$. 
\end{enumerate}
The loss computes how far away a given partition is from our desired objectives. The loss has two factors: (i) the \emph{quality cost}, which measures how bad on average a candidate set is for a query, and (ii) the \emph{computational cost}, which measures how far away the partition is from being a balanced one.

We define the terms used in the loss formulation in Table \ref{tab:notation_table}:

\begin{table}[ht]
\centering
\begin{tabular}[t]{|c| p{0.7\linewidth} |}
\hline
   \textbf{Notation} & \textbf{Meaning}\\
\hline
   $X \in \mathbb{R}^d$ & The $d$ dimensional dataset to be partitioned\\
\hline
   $Q$ & Set of queries $\{q_1, q_2,.. \}$, not necessarily present in $X$ \\ 
\hline
   $R$ & A partition that divides $X$ into $m$ bins\\
\hline
 $N_{k^\prime}(p)$ & set of true $k^\prime$-nearest neighbors of point $p$ from $X$\\
 \hline
 
     $R(p)$ & the most likely bin \tablefootnote{Note that our model returns the probability distribution of a point being in different bins of the given partition. In order to formulate the loss during model training, we only consider the most likely bin the model assigns to an input point.} in $R$ that might contain $p$\\
\hline
    $C(p)$ & Candidate set of $p$\\

\hline
\end{tabular}
\caption{Notations used}

\label{tab:notation_table}
\end{table}

\revise{In $N_{k^\prime}(p)$, $p \in \mathbb{R}^d$ can either be a query point not present in $X$, or a data point in $X$. Note that the $k^\prime$-NN matrix we defined earlier helps us to quickly retrieve $N_{k^\prime}(p_i)$ for any point $p_i$ by simply indexing into the $i$th row of the $k^\prime$-NN matrix. }

\revise{For a given partition $R$, $C(p)$ is the set of all points in $X$ that are present the bin $R(p)$. Therefore, for a point $p$, $C(p)$ denotes its \emph{candidate set}.}

\revise{Finally, $Q$ denotes the set of all query points, where points in $Q$ are not necessarily present in $X$.}

We can now define the quality cost and the computation cost of $R$ as follows:

\begin{itemize}
    \item  
        The quality cost of $R$, $U(R)$, can be defined as: \begin{equation}
            \label{quality_eqn}
            U(R) = \sum_{q \in Q} \sum_{p \in N_{k^\prime}(q)}  \mathbbm{1}_{R(p) \neq R(q)}
        \end{equation}
        \begin{itemize}
            \item \revise{Where $\mathbbm{1}$ is the indicator function. The factor $\mathbbm{1}_{R(p) \neq R(q)}$ can otherwise be expressed as:}
            \begin{equation}
                \mathbbm{1}_{R(p) \neq R(q)} = 
                \begin{cases}
                  1, &  \text{if }  R(p) \neq R(q) \\
                  0, & \text{otherwise}
                \end{cases}
            \end{equation}
            where $R(p) \neq R(q)$ if the bin in $R$ that contains $p$ is not the same as the bin that contains $q$. 
        \end{itemize}
    
    \item The average computation cost of $R$, $S(R)$, can be determined by taking the mean of the candidate set sizes of all the query points:
    \begin{equation}
        \label{comp_eqn}
        S(R) = \underset{q \in Q}{\text{mean }} |C(q)|
    \end{equation}
    
\end{itemize}

To create a partition that serves as an efficient index for searching the $k$ nearest neighbors, we need to find $R$ that minimizes both $U(R)$ and $C(R)$. Mathematically,

\begin{equation}
    \label{loss-formula}
   R_{\text{optimal}} = \min_{R} \{U(R) + \eta . S(R)\}
\end{equation}

where $\eta$ is a balance parameter that trades off between the two factors of the cost.

We can implement our loss function using any standard modern machine learning library that supports tensor operations with automatic differentiation, which will allow the framework to compute the gradients of our loss function with respect to the parameters of any machine learning model without explicitly formulating them.

\textbf{Computing quality cost}: For simplicity, let us assume that any data point $p_i$ can be a query. Now, we show how to compute $U(R)$ for a single data point, $p_i \in \mathbb{R}^d$, in $X$.

First, we input $p_i$ into the model $M$, to get $b_i$ as follows.
\begin{equation}
    M(p_i) = b_i =  \begin{pmatrix}
    b_i^1 & b_i^2 & ... &  b_i^m
    \end{pmatrix}
\end{equation}
Here $M(p_i)$ is the model's output for the point $p_i$, and $b_i^j$ is the probability of point $i$ being assigned to bin $j$.

We now determine to which bin $p_i$ should be assigned if the partition is optimal. To do this, we use the $k^\prime$-NN matrix to quickly retrieve $N_{k^\prime}(p_i)$, the set of true $k^\prime$-nearest neighbors of $p_i$ from $X$, as:
\begin{equation}
    N_{k^\prime}(p_i) = \begin{pmatrix}
                    \hat{p}_1 & \hat{p}_2 & ... & \hat{p}_{k^\prime}
                \end{pmatrix}
\end{equation}

Here $\hat{p}_j$ is the $j$th nearest neighbor of $p_i$ in $X$.


We pass all the points in $N_{k^\prime}(p_i)$ through the model to get the model's outputs for the ${k^\prime}$-nearest neighbors of $p_i$.
\begin{equation}
    M\{N_{k^\prime}(p_i)\} = \begin{matrix}
\hat{b}_1 \\
...\\
\hat{b}_{k^\prime}
\end{matrix}
\begin{pmatrix}
\hat{b}_1^1 & \hat{b}_1^2 & ... &  \hat{b}_1^m\\
... \\
\hat{b}_{k^\prime}^1 & \hat{b}_{k^\prime}^2 & ... & \hat{b}_{k^\prime}^m
\end{pmatrix}
\end{equation}

Here, $\hat{b}_j$ is the model's output for the $\hat{p}_j$.

\revise{Next, we determine the distribution of the points in $N_{k^\prime}(p_i)$ among the available bins. To do this, we take the proportion of points assigned to each bin from $M\{N_{k^\prime}(p_i)\}$ to get the following.}

\begin{equation}
    B_{k^\prime}(p_i) = \begin{pmatrix}
                        \hat{B}_1 & \hat{B}_2 & ... & \hat{B}_{m}
                    \end{pmatrix}
\end{equation}

\revise{where, $B_{k^\prime}(p_i)$ lists the proportion of points among the $k^\prime$-NNs of $p_i$ that  belong to each bin.}

\revise{Ideally, we want the model output for $p_i$ to indicate the distribution of its nearest neighbors over all the bins. Therefore, we take $B_{k^\prime}(p_i)$ as the ground truth labels for the point $p_i$ and compute $p_i$'s quality loss as the \textbf{cross entropy loss} between $B_{k^\prime}(p_i)$ and $M(p_i)$:}
\begin{equation}
    \label{qr_per_point}
    U(R) \text{ for } p_i = \text{cross\_entropy\_loss}(b_i, B_{k^\prime}(p_i))
\end{equation}

Finally, to compute $U(R)$ for the entire dataset $X$, we calculate $U(R)$ using Equation ~\ref{qr_per_point} for every point in $X$ and then take the average.

\textbf{Computational cost}: For determining the computation cost factor of the loss function, $S(R)$, we need the model's output on all the points in the dataset $X$. We pass the entire $X$ through the model, $M$, to get the following output as $M(X)$.
\begin{equation}
\label{model_outputs}
    \begin{pmatrix}
    b_1^1 & b_1^2 & ... &  b_1^m\\
    ... \\
    b_n^1 & b_n^2 & ... & b_n^m
    \end{pmatrix}
\end{equation}

Here, $b_i^j$ is the probability that the model assigned point $i$ to bin $j$.

Our target is to make the model evenly distribute the $n$ points in $X$ among all the $m$ available bins. Therefore, we ideally want each bin to contain $n/m$ points. In the model outputs, $M(X)$, in Equation ~\ref{model_outputs}, each $i$th row denotes the model outputs for the $i$th point in $X$, and the $j$th column denotes the probabilities of assigning each of the $i$ points to the $j$th bin.

\revise{To ensure an even distribution of points between the available bins, we want all the $n$ points in the dataset to be assigned to the $m$ available bins evenly, such that each bin has approximately $n/m$ points assigned to it. For each query point, $q$, our model outputs a probability distribution over the available bins for assigning $q$. We assign $q$ to the bin with the highest probability from this distribution. Therefore, for a balanced partition, we want each column to only have $n/m$ \emph{high} values, since the $i$th \emph{high} probability value in the $j$th column corresponds to point $i$ being assigned to the $j$th bin. To that end, we \emph{filter} the highest $n/m$ probability values by selecting the highest $n/m$ values in each column of the output matrix to get a \emph{window}, $w$, of high probability values:}
\begin{equation}
    \label{window}
    \begin{split}
    w &= \text{max $n/m$ values across each column of } M(X)\\ 
       &= \begin{pmatrix}
            b_1^1 & b_1^2 & ... &  b_1^m\\
            ... \\
            b_{n/m}^1 & b_{n/m}^2 & ... & b_{n/m}^m
        \end{pmatrix} 
    \end{split}
\end{equation}

To calculate $S(R)$, we sum all the entries in the window, $w$, from Equation ~\ref{window} and negate it:
\begin{equation}
    S(R) = -\sum{\text{w}}
\end{equation}

Minimizing $S(R)$ leads to higher values in the $n/m$ window, creating a more balanced partition.

\textbf{Caveats}: In the operations detailed above, we calculate the loss using only the \textbf{data points} in $X$, even though our loss formulation in Equations ~\ref{quality_eqn} and ~\ref{comp_eqn} requires a set of \textbf{query points}. In our formulation, we assume that the query points follow the same distribution as the data points in $X$. Therefore, we can use only the points in $X$ to compute the loss.

Another caveat of our loss is that we can only calculate it over a batch of input points and not for individual data points like in other loss functions typically used in machine learning (We calculate $S(R)$ over the entire batch of points). We need a batch of points to compute $S(R)$ because the model cannot learn anything about the underlying distribution of $X$ from a single data point. As a result, we need special care when using mini-batches for model training.

\textbf{Batching}: So far, we assume that the output matrix of the whole dataset is available to us for calculating the loss value. In practice, the output matrix of the entire dataset may not fit in GPU or CPU memory during model training. In this case, we can approximate the data distribution by randomly sampling a smaller batch of points from the dataset for each iteration of the training loop. As long as our sampling technique is uniform (i.e., we choose every point in $X$ for a particular mini-batch with equal probability), the sampled mini-batch will have roughly the same distribution of points as $X$. Our experiments show that sampling even just $\approx 4 \%$ of the dataset per mini-batch leads to relatively high-quality learned partitions.

\begin{algorithm}[t]
  \caption{Offline Phase - Train model to create space partitioning index}
  \label{offline_algo}
    \textbf{Input}: Dataset $X \in \mathbb{R}^d$, nearest neighbors to use  $k^\prime > 0$ , number of bins $m$, Distance function $D$
    \begin{enumerate}
      \item \label{offline_1} Create a $k^\prime$-NN matrix by computing pairwise distances using $D$ between all points in $X$, then storing indices of true $k^\prime$ nearest neighbors of each point.
      \item \label{offline_2} Train a machine learning model $M$ with the loss function defined in ~\ref{loss}. This model jointly learns a partition of $X$ \textbf{and} learns to classify new points to assign queries into bins.
      \item \label{offline_3} Run inference on all points in $X$ to form a partition $R$ of $X$. Store the point indices to keep track of the points in $X$ assigned to each bin in a lookup table.
    \end{enumerate}

\end{algorithm}

\subsubsection{Training the Model}

Algorithm ~\ref{offline_algo} outlines the whole learning process. We detail the algorithm steps below.

In Step ~\ref{offline_1} we create the $k^\prime$-NN matrix using a given distance measure $D$. Then, in Step ~\ref{offline_2}, we use the points in $X$, the $k^\prime$-NN matrix, and the loss function defined above to train a model to create a partition of the dataset $X$ with $n$ points $\in \mathbb{R}^d$, dividing it into a predetermined number (say $m$) of bins. We use the machine learning model in this setting to output a probability distribution over the bins assigned to $q$. 

We want our model to generalize well to query points $(q \in \mathbb{R}^d)$ outside of $X$ (i.e., queryWedel has never seen during training). Therefore, we have to cluster the dataset $X$ into $m$ bins and also partition the entire $\mathbb{R}^d$ for the range occupied by the dataset. Neural networks are suitable for this task. They can learn complex decision boundaries optimized for a specific dataset and use regularization techniques to prevent overfitting on the training data. We learn the partition by minimizing the loss function defined in Section ~\ref{loss}. 

After the model training is complete, in Step ~\ref{offline_3}, we pass the entire dataset of points ($X$) through the model to obtain the learned partition of the dataset $X$. In the online phase, we need to quickly retrieve all the points in $X$ belonging to a particular bin. To speed up this retrieval, we store the indices of the points in $X$ assigned to each bin in a lookup table.

\subsection{The Online Phase}

Once the system trains the model and creates the lookup table outlined in the previous section, it is ready to answer queries in the online phase. Algorithm ~\ref{online_algo} outlines the online phase.

In Step ~\ref{online_1}, we pass the given query point $q$ through the model to get $M(q)$, a probability distribution over assigned bins of $q$. In step ~\ref{online_2}, $M(q)$ is used to determine the set of bins $b_q$ the query point might belong to. Then, using the lookup table created in the offline phase, we retrieve all the points in $X$ that also belong to the bins in $b_q$ to form the \emph{candidate set} of points, $C(q)$. Finally, in Step ~\ref{online_3}, we search through the points in $C(q)$ to return the $k$-Nearest Neighbors of $q$. Hence, we reduce the search space from the entire dataset to just $C$. 

Instead of searching in just one bin, we use the probability distribution output by the model to search in the $m^\prime$ most probable bins. This way, we trade-off higher nearest neighbors accuracy (since we are more likely to find neighbors close to $q$ simply by searching through more nearby points) at the cost of higher search time (since we need to search through a larger candidate set).

\begin{algorithm}[h]
    \caption{Online Phase: Return the k-nearest neighbors for a query point}
    \label{online_algo}
    \textbf{Input}: Query Point $q \in \mathbb{R}^d$, number of bins to search $m^\prime$, number of nearest neighbors to return $k$, Distance function $D$, Trained model $M$.
     \begin{enumerate}
        \item \label{online_1} Run inference on point $q$ by computing $M(q)$
    \item \label{online_2}From $M(q)$, for the most probable $m^\prime$ assigned bins $b_q = \{b_1, b_2, ..., b_{m^\prime} \}$, retrieve all points from $X$ that are assigned to any of $b_q$, using the lookup table from Step \ref{offline_3} in Algorithm ~\ref{offline_algo}, to form the \emph{Candidate Set} $(C)$ 
        \item \label{online_3}For all points in $C$, compute $D(q, p_i)$, and return the $k$ most similar points to the query.
      \end{enumerate}
\end{algorithm}

\subsection{Optimizations}

\label{optimizations}

In this section, we propose two additional components: (i) A boosting method that uses an ensemble of models to create multiple partitions, and ii) a hierarchical partitioning strategy that recursively divides the dataset to get finer dataspace partitions. 

\subsubsection{Ensembling} \label{ensembling}

\begin{figure*}[h]
	
    \includegraphics[width=0.7\textwidth]{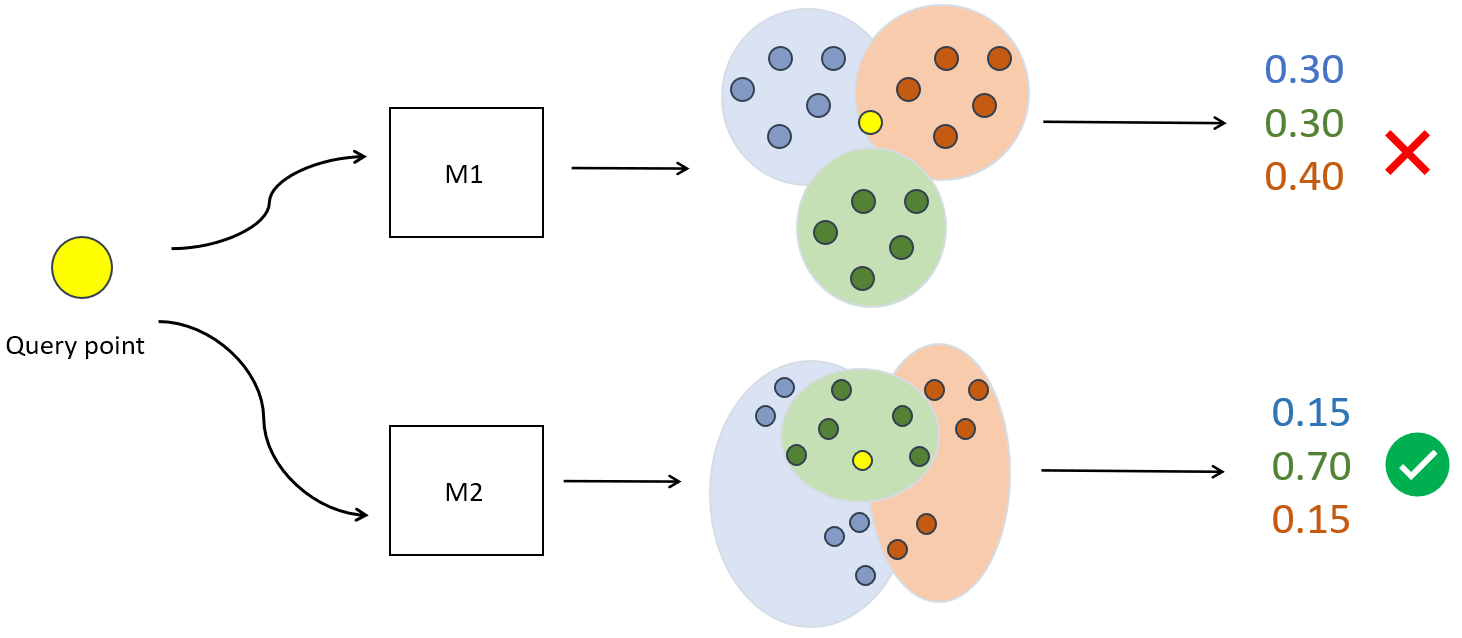}
    \captionsetup{justification=centering}
    \caption{Ensembling with two models. Here, Model 2 (M2) performs better with the yellow query point, resulting in the second model outputting a higher confidence value.}
    \label{ensemble_fig}
\end{figure*}

In applications where high $k$-NN accuracy is crucial, we can boost the accuracy by training multiple models sequentially, with each model generating a different partition for the same dataset. We call this approach \emph{ensembling}, where we create an \emph{ensemble of models}. Ensembling allows us to create a set of \emph{complementary partitions} for a single dataset. The intuition behind ensembling is that different models can \emph{specialize} in different regions of the data space. Working together, these models can increase the quality of candidate sets generated for any query point. Figure ~\ref{ensemble_fig} illustrates this intuition. 

\begin{algorithm}
  \caption{Ensembling}
  \label{ensembling_alg}
    \textbf{Input}: Dataset $X \in \mathbb{R}^d$ containing $n$ points, Initial input weights \revise{$W_1 = \{ w_{1}^{1}, w_{2}^{1}, ..., w_{n}^{1}\}$}, Number of models in ensemble $e$
    \begin{enumerate}
      \item \textbf{for} $j \in 1, 2, ...e$ \textbf{do}:
      \begin{enumerate}
          \item Train model $m_j$ to learn partition $r_j$, using weights $W_{j}$, by modifying the quality cost of the loss function:
          \begin{equation}
            \label{quality_ensemble_eqn}
          U(r_j) = \sum_{i=1}^{n} q_i.w_i^{\revise{j}} \sum_{p \in N_{k^\prime}(q_i)}  \mathbbm{1}_{r_{j}(p) \neq r_{j}(q_i)} 
      \end{equation}
      
      \item Obtain new weights for use in the next model: 
      $$ w_{i}^{\revise{j+1}} =  \sum_{p \in N_{k^\prime}(q_i)} \mathbbm{1}_{R(p) \neq R(q_{\revise{i}})}  $$
      
      $$ w_{i}^{\revise{j+1}} = w_{i}^{\revise{j+1}}. w_{i}^{j} $$
      \end{enumerate}
    \end{enumerate}

\end{algorithm}

\begin{algorithm}
  \caption{Querying with ensembling}
  \label{query_ensemble_alg}
    \textbf{Input}: Query point $q$, Ensemble of trained models $(M_1, M_2, ..., M_e)$
    \begin{enumerate}
      \item Run inference on the query point $q$ on all the models $(M_1, M_2, ..., M_e)$ in the ensemble to get corresponding bin assignments of each model.
      \item Each model, $M_i$, returns a candidate set, $c_i$, 
      $$ C = \{c_1, c_2, ..., c_e\} $$
      
      \item Take each model's highest probability as its confidence value, $\sigma_i$: 
      $$ S = \{\sigma_1, \sigma_2, ..., \sigma_e \}$$
      \item the best candidate set is the one with the highest confidence score:
      $$ c_{best} = C\left[{arg\,max_{S}} \right] $$
      
      \item search through the items as before on only the best candidate set to return the nearest neighbors of $q$
    \end{enumerate}

\end{algorithm}

\revise{Our ensembling algorithm is based on AdaBoost \cite{schapire2013explaining}. However, unlike AdaBoost, instead of training many \emph{weak learners}, we use this boosting formulation to create many complementary partitions, to improve the quality of the generated candidate set. \emph{Boosted Search Forest} \cite{li_learning_2011} used this concept in a similar fashion}.

To create an ensemble of models, we first assign weights to each point in $X$. We update the quality cost factor of the loss function as in Equation ~\ref{quality_ensemble_eqn} in Algorithm ~\ref{ensembling_alg} to incorporate these weights. We train the different models in the ensemble sequentially. We assign equal weights to all the data points for training the first model. After training the first model, we use the trained model to obtain new input weights for the second model. We can then train the second model using the new input weights and so on. In Algorithm ~\ref{ensembling_alg}, $w_{i}^{j}$ represents the $ith$ data point's weight for the $jth$ model in the ensemble.

Intuitively, each model tries to optimize its partition to perform better for points with which \emph{all} the previous models performed poorly. Each model in the ensemble will tune its partition to give more importance to "difficult" points (i.e., points with a high weight value) since they contribute more to the quality factor of the loss. We ensure that the weights of the following models in the ensemble only try to optimize for the points in which previous models could not do well by multiplying the weights of all points with the weights of the previous models. Multiplying the weights like this ensures that only points with high weights for all previous models will have high weights for the next model.

In the online phase, we pass the query point $q$ through all the models in the ensemble. Since each model $M_i$ returns a probability distribution over assigned bins, we can return the highest probability as the confidence value of $M_i$. Then, we select the candidate set corresponding to the model with the highest confidence value as the output candidate set of the ensemble. Algorithm \ref{query_ensemble_alg} outlines the querying process.

\subsubsection{Hierarchical Partitioning}

\begin{figure}
    \centering
    \includegraphics[width=0.4\textwidth]{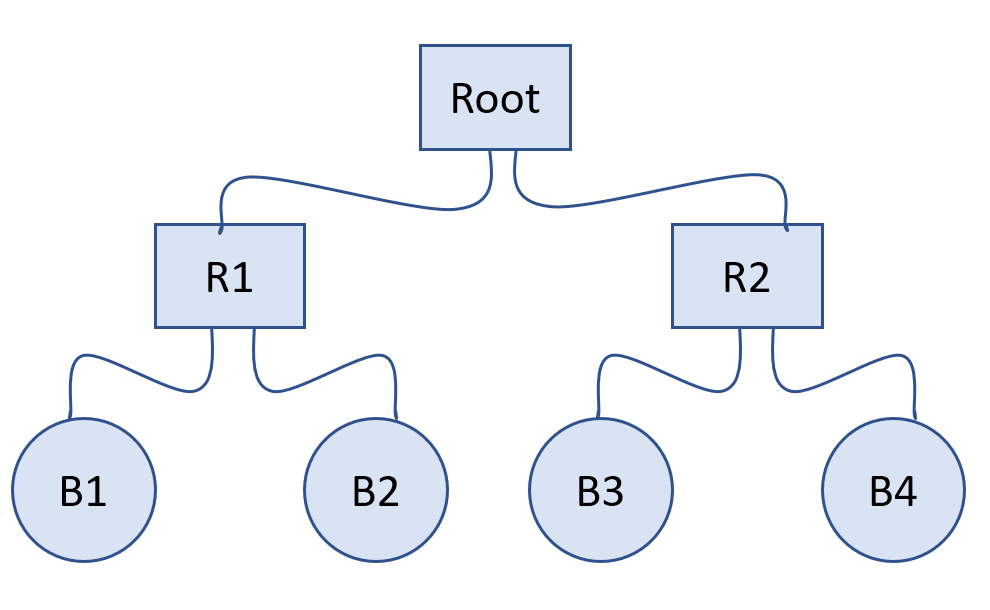}
    \captionsetup{justification=centering}
    \caption{Dividing a dataset hierarchically with three models (one root model and two leaf models), finally resulting in a partition with four bins}
    \label{model_tree}
\end{figure}

When the number of required bins $m$ is large, training can become difficult as we attempt to partition a large dataset into many bins in a single pass. In order to make training more efficient, we can recursively partition the dataset into $m_1$ bins at the first level, then subdivide each of those bins into $m_2$ bins at the second level, and so on, resulting in a total of $m_1 \cdot m_2 \cdot .... \cdot m_l$ bins for $l$ level-partitioning. This is illustrated in Figure ~\ref{model_tree}. 

For a query point $q$, we pass $q$ from the top of the tree down to the leaves. We multiply the assigned probabilities of each model down the tree to obtain the final probability of assigning $q$ to each of the bins in the leaves.  
Hierarchical partitioning allows us to simplify the learning process for each model. Further, each model can have fewer parameters and be simpler since each model's learning task is more straightforward. As a result, we can often train a tree of models that takes up lesser total memory than a single large and complex model needed to partition the same dataset in a single pass. 

\subsection{Time Complexity Analysis}

\revise{The online phase of our algorithm is sublinear as we do not have to traverse the entire dataset to find a query's k-NNs. For a given query point $q$, our algorithm follows two steps to find $q$'s k-NNs. First, we feed $q$ to our model to find the associated bins of $q$ and thus its candidate set. Second, we traverse the candidate set to find $q$'s nearest neighbors (by brute-force search). The first task is of order $d$, the dimensionality of $q$, since the input layer of the trained model takes $d$ values for multiplication. The second task is of order $cd$, where $c$ is the largest candidate set size, since we need to traverse the entire candidate set to find $q$'s k-NNs. Thus, finding k-NNs of a single query point $q$ using our approach is an operation of order $O(cd + d)$.}

\section{Experiments}\label{experiments}

We present detailed experimental evaluations of our proposed approach and compare the results with the state-of-the-art baselines using several real datasets. We first discuss the experimental settings that include datasets, baselines, performance metrics, and parameters of the experiments. We then discuss the implementation details of the algorithm and present our experimental evaluation. Finally, we compare our space-partitioning performance with that of common clustering methods.

\subsection{Experimental Settings}
Here, we discuss the datasets, state-of-the-art baseline approaches, and different parameters of our experiments.

\subsubsection{Datasets}
For our experimental benchmarks, we used two standard ANN benchmark datasets ~\cite{aumuller_ann-benchmarks_2018}: 
\begin{itemize} 
    \item \textbf{SIFT}: 1M data points, each having 128 dimensions
    \item \textbf{MNIST}: 60k data points, each having 784 dimensions
    
\end{itemize}

Both datasets come with 10k query points that are not present in the training dataset. We choose these datasets as they encompass both aspects of large-scale datasets: a high number of points (SIFT has 1M points), and high dimensionality (MNIST has 784 dimensions), with data taken from real-world applications.

\subsubsection{Baselines}

We compare our approach with several space partitioning baselines, outlined in Section ~\ref{implementation_details}. Notably, we compare with the state-of-the-art Neural LSH ~\cite{dong_learning_2020} and K-means clustering. Neural LSH~\cite{dong_learning_2020} is currently the best-performing deep learning based space-partitioning approach. On the other hand, K-means clustering is a well-known technique used in many production systems for partitioning the dataset before ANN search or other processing. For both baselines, we use the same codebase and settings found in the Neural LSH~\cite{dong_learning_2020} paper: ~\url{https://github.com/twistedcubic/learn-to-hash}. 
\revise{To demonstrate how our partitioning strategy can enhance the performance of the state-of-the-art non-learning ANNS techniques, we incorporate our method with ScaNN and compare the performance with vanilla ScaNN~\cite{guo_accelerating_2020}, HNSW~\cite{malkov_efficient_2018}, and FAISS~\cite{johnson_billion-scale_2017}.}

\subsubsection{Performance metrics}
To evaluate the effectiveness of the baseline approaches, we compare and evaluate the trade-offs between two key metrics:
\begin{enumerate}
    \item The $k$-NN accuracy: The fraction of the true $k$-Nearest Neighbors ($k$-NN) that are present among the $k$ returned points by the algorithm.
    \item The size of the candidate set: The number of points in the candidate set $C$ represents the query processing time, as we need to search through all the points in $C$ to return the $k$-NN.
\end{enumerate}
In general, more candidates present in the candidate set for any partitioning (or clustering) algorithm lead to a larger $k$-NN accuracy. 

\subsubsection{Parameters}

Our algorithm exposes a lot of tuneable parameters for the user to optimize the framework to their specific application needs. Changing each of these parameters affects a different part of the model. These parameters include:

\begin{enumerate}
    \item Integer $k^\prime$: This value specifies the number of nearest neighbors to consider when building the $k^\prime$-NN matrix in the offline phase. Setting a larger $k^\prime$ provides more information to the model and loss during training at the cost of requiring more memory during training. However, setting $k^\prime$ too high would result in far-away points becoming nearest neighbors for many data points. We found that setting $k^\prime$ to 10 creates sufficiently good dataset partitions while using less memory during training. Also, setting larger values of $k^\prime$ does not appreciably increase the quality of the created partitions. 
    
    \item Integer $m$, number of bins to split the dataset into: $m$ affects how finely the model splits the dataset during training and, in turn, how "difficult" the problem is for the neural network. Setting $m$ to 16 for a 1M sized dataset, for instance, means that the dataset will be almost evenly split among $16$ bins, resulting in about $1M / 16 = 62500$ points per bin. On the other hand, setting $m$ to 256 for a 1M sized dataset partitions the dataset into 256 bins, with each bin having $1M / 256 \approx 3900$ points.
    
    \item Integer $e$, number of models in the ensemble: $e$ denotes the number of models to train for a single dataset. Each of the $e$ models describes a different partition of the dataset. Since each model optimizes for the poorly placed points in all previous partitions, having more models increases $k$-NN accuracy for the same candidate set size. Also, having a larger $e$ means that each model can be simpler and can afford to learn simpler (might not be high-quality) partitions (using a neural network with fewer parameters). Learning simpler models does not sacrifice partitioning quality since the greater number of models in the ensemble can boost the quality of the returned candidate set of the individually weak models. However, a larger $e$ comes at the cost of longer training times (since each of the $e$ models trains sequentially) and higher memory usage (since each of the models must be stored, along with their individual lookup tables).
    
    \item Model Complexity: In our proposed framework, we can use any machine learning model architecture as $M$, the model used to learn the partitions. For instance, increasing the number/size of the hidden layers or using a more complex architecture (such as replacing a linear model with a neural network) results in better-learned partitions. However, more complex models require longer training times and more memory to store the larger models. We demonstrate this by training two different model architectures, a \textbf{neural network} and a \textbf{logistic regression} model, and presenting their results in Sections ~\ref{neural-methods} and ~\ref{tree-methods}.

    \item \revise{$\eta$: The balance parameter in the loss (Equation \ref{loss-formula}). This value quantifies the trade-off between the two factors of the loss function. Increasing $\eta$ makes the partition more balanced, but a value of $\eta$ too high makes it more difficult for the model to optimize the quality cost factor of the loss function}. We tuned $\eta$ and set it to the lowest value, resulting in a balanced partition. We mentioned the specific values of $\eta$ used in Table~\ref{table:train-times}. 
    
\end{enumerate}

\subsection{Implementation Details}
\label{implementation_details}
We demonstrate our partitioning performance with two different model architectures:
\begin{itemize}
    \item \textbf{Neural Networks}: Here, we used a small neural network with one input layer and one hidden layer containing 128 parameters. Each network layer consists of a fully connected layer, and batch normalization ~\cite{ioffe_batch_2015}, followed by ReLU activations. The final layer is an output layer containing $m$ output nodes followed by a softmax layer, where $m$ is the number of bins in the partition. To reduce overfitting and to generalize well to unseen queries, we use dropout ~\cite{srivastava_dropout_2014} with a probability of 0.1 during training. We train each neural network for about 100 epochs. We compare this model's performance with baselines K-means clustering and Neural LSH ~\cite{dong_learning_2020}. We also include results for the data oblivious Cross-polytope LSH ~\cite{andoni_practical_2015} to show improvements in the performance of learning methods over non-learning methods. 
   
    \item \textbf{Logistic Regression}: Here, we used a simple logistic regression model to divide the dataset into two bins at each level recursively to form a partitioning tree. Each model in the tree has two output nodes in the final layer, followed by a softmax layer to output a probability distribution over two bins. We trained each logistic regression model for less than 50 epochs. We compare this model's performance with other tree-based partitioning methods that recursively split the dataset using hyperplanes: \emph{Regression LSH} ~\cite{dong_learning_2020} (A variant of Neural LSH that uses logistic regression instead of neural networks), 2-means tree, PCA trees ~\cite{sproull_refinements_1991, forsyth_what_2008, abdullah_spectral_2014}, Random Projection trees ~\cite{dasgupta_randomized_2013} , Learned KD-tree ~\cite{cayton_learning_2007}, and Boosted search forest ~\cite{li_learning_2011}.

\end{itemize}

The model weights were initialized for both architectures with Glorot initialization ~\cite{glorot_understanding_2010}. We trained both types of models using the Adam optimizer ~\cite{kingma_adam_2017}. To show the performance improvements of ensembling,  we used an ensemble of methods to boost the retrieval performance of the neural network architecture in our experiments. 

In our experiments, we use the same number of bins for all the methods to evaluate our approach's representative performance. We use PyTorch ~\cite{paszke_pytorch_2019} to implement our algorithms.

\subsection{Training Efficiency}

\revise{We trained our models on a hosted runtime with a single-core hyperthreaded Xeon processor, 12GB RAM, and a Tesla K80 GPU with 12GB GDDR5 VRAM. Training multiple models in an ensemble with million-sized datasets takes less than an hour, significantly lower than the several hours of preprocessing time needed for Neural LSH. We highlight the different training times for different specifications in Table \ref{table:train-times}. The training times mentioned in Table \ref{table:train-times} are the total times needed to train three base models in the ensemble while keeping GPU usage under \textbf{6GB}.}

\revise{We also need significantly fewer parameters on even the largest model sizes to beat Neural LSH's partitioning performance when dividing the dataset into 256 bins. We highlight this in Table ~\ref{table:params}.}

\begin{table}[H]
\begin{center}
\begin{tabular}{ c | c c c}
 & Neural LSH & Ours & K-Means \\ 
 \hline
 No. of bins & 256\\  
 Total parameters &  729k & \revise{183k} & 33k \\ 
 Hidden layer size & 512 & 128 & -
\end{tabular}
\captionsetup{justification=centering}
\caption{Approximate number of learnable parameters of selected space-partitioning methods when dividing SIFT into 256 bins.}
\label{table:params}
\end{center}
\end{table}

\begin{table}[H]
\begin{center}
\begin{tabular}{ c c | c c }
 Dataset & No. of bins & \textbf{Training time (minutes)} & \textbf{Value of $\eta$}\\ 
 \hline
 MNIST & 16 & 2min & 7\\  
 MNIST & 256 & 12min & 30\\ 
 SIFT & 16 & 6min & 7\\  
 SIFT & 256 & 40min & 10\\ 
\end{tabular}
\captionsetup{justification=centering}
\caption{\revise{Comparing our method's approximate offline training times and $\eta$ values with different configurations. 
}}
\label{table:train-times}
\end{center}
\end{table}


\subsection{Performance Evaluation}

We evaluate the performance of our method by comparing it with space-partitioning methods using a neural network model and tree-based methods using a logistic regression model.

We generate each of the graphs shown by successively searching in more of the most probable bins returned by the algorithms. We systematically note the k-NN accuracies with increasing candidate set size, $|C|$.  

\subsubsection{Comparing with space-partitioning methods}
\label{neural-methods}
\begin{figure*}[h]
     \centering
     \begin{subfigure}[b]{0.45\textwidth}
         \centering
         \includegraphics[width=\textwidth]{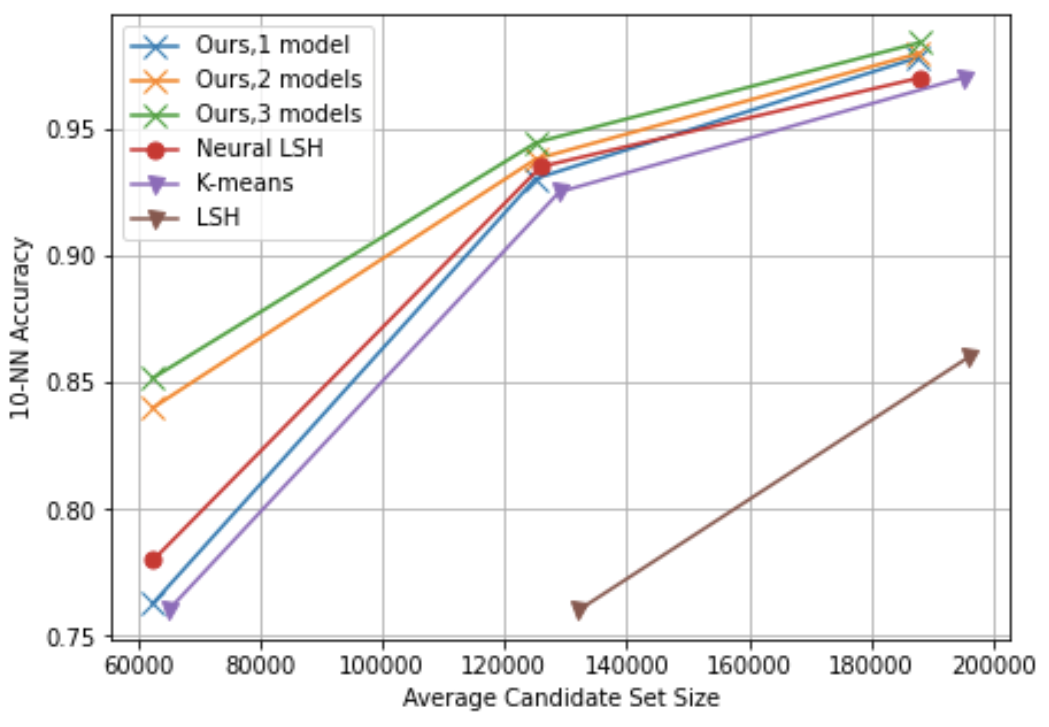}
         \caption{SIFT, 16 bins}
         \label{sift-16}
     \end{subfigure}
    \hspace{5mm}%
     \begin{subfigure}[b]{0.45\textwidth}
         \centering
         \includegraphics[width=\textwidth]{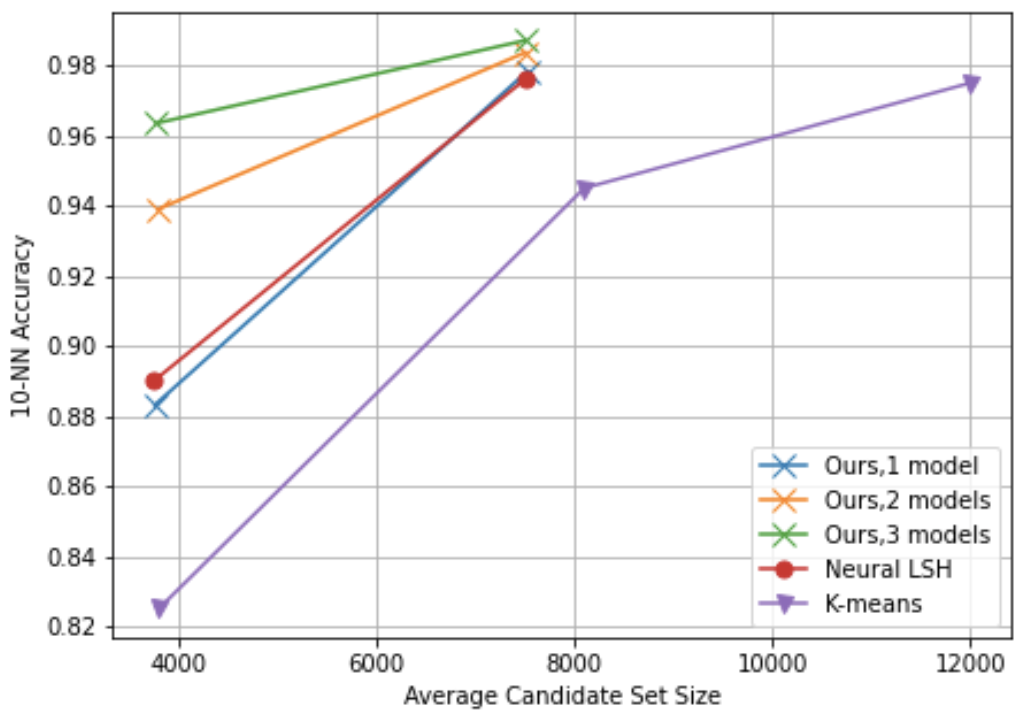}
         \caption{MNIST, 16 bins}
         \label{mnist-16}
     \end{subfigure}
    \hspace{5mm}%
     \begin{subfigure}[b]{0.45\textwidth}
        
         \centering
         \includegraphics[width=\textwidth]{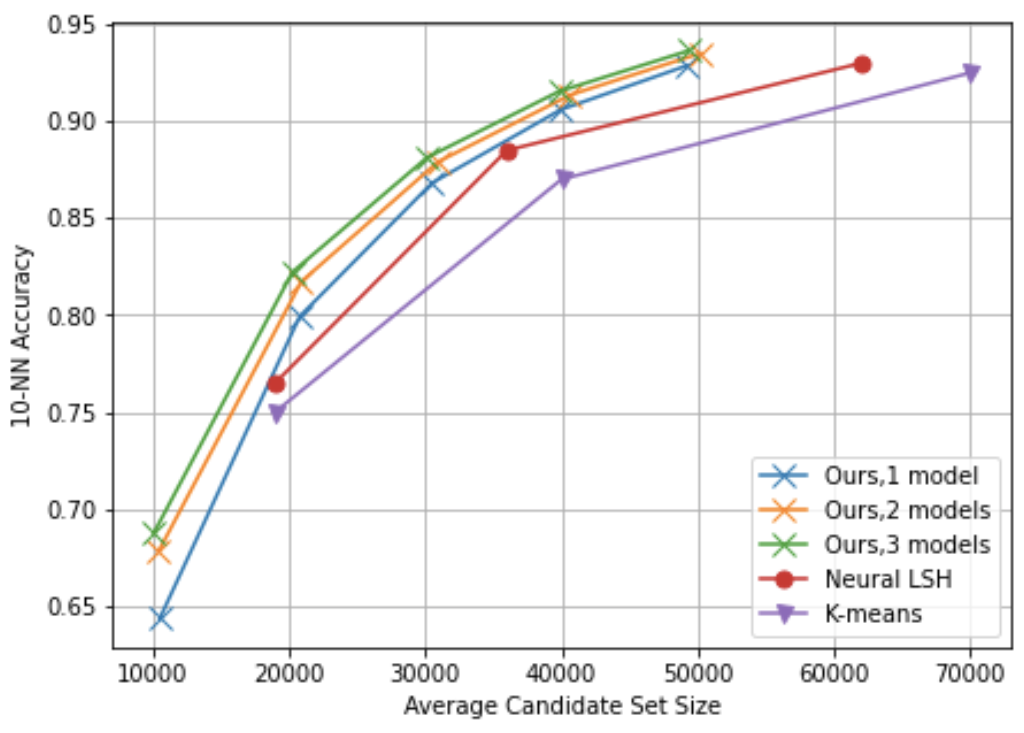}
         \caption{SIFT, 256 bins}
         \label{sift-256}
     \end{subfigure}
    \hspace{5mm}%
     \begin{subfigure}[b]{0.45\textwidth}
         \centering
         \includegraphics[width=\textwidth]{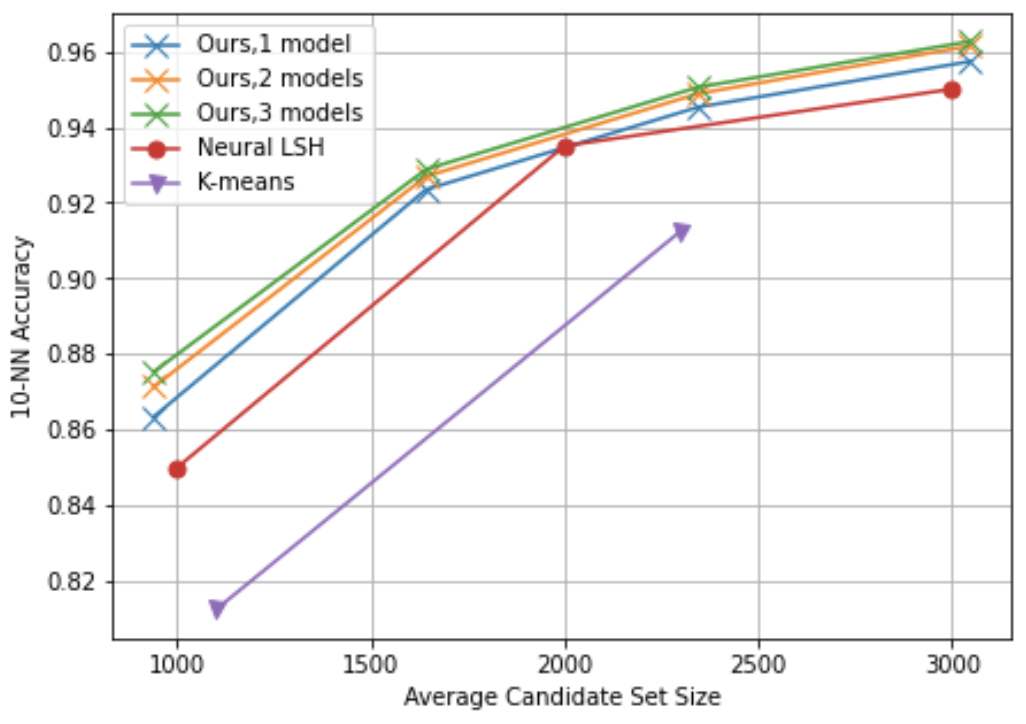}
          \captionsetup{justification=centering}
         \caption{MNIST, 256 bins}
         \label{mnist-256}
     \end{subfigure}
     \captionsetup{justification=centering}
    \caption{Comparing our method with space-partitioning baselines. X-axis: number of candidates retrieved in the candidate set. Y axis: 10-NN accuracy (Up and to the left is better). Our method uses an ensemble of 3 models to boost performance.}
    \label{fig:graphs}
\end{figure*}

Here, we present the performance evaluation of our proposed approach, using a \textbf{neural network} as the learning model. Figure ~\ref{fig:graphs} shows the comparison between our method and the selected baselines: Neural LSH, K-means, and Cross polytope LSH. We test with 16 and 256 bins for all the baselines for the experiments to show the trade-off between candidate set sizes and 10-NN accuracies. \revise{We use hierarchical partitioning when dividing the dataset into 256 bins, first splitting into 16 bins and then sub-splitting each bin into 16 more bins.} Splitting the dataset into a greater number of bins allows us to control the candidate set size, $|C|$, more finely because searching each additional bin of points increases $|C|$ by a smaller amount. This leads to more points in the graph in Figures ~\ref{sift-256} and ~\ref{mnist-256}. 

\revise{We see that our model performs better than Neural LSH even using just one base model in the ensemble when partitioning the dataset into 256 bins (in figures ~\ref{sift-256} and ~\ref{mnist-256}). Partitioning the dataset into a larger number of bins is an expected configuration. It leads to greater k-NN accuracy in the online phase with smaller candidate set sizes at the expense of longer training times and larger models.}

\revise{As for partitioning into 16 bins, we see almost similar partitioning performance compared to Neural LSH with both datasets in Figure ~\ref{fig:graphs} when we do not use any ensembling and train just one model. The similarity in k-NN retrieval performance suggests that our model learns similar partitions to Neural LSH without using any graph partitioning algorithm in an unsupervised setting and uses significantly less time. When using more than one model in an ensemble, we see up to about $10\%$ improvement in k-NN accuracy using three models (Figure ~\ref{fig:graphs}). }

Table ~\ref{table-compare} shows the relative decrease in our method's average candidate set sizes compared to Neural LSH and K-means when dividing the SIFT dataset into 16 bins and maintaining a 10-NN accuracy of 85\%. The smaller candidate set sizes speed up ANNS proportionately as we have to search through a smaller number of points to attain the same 10-NN accuracy.

The experiments show that while Neural LSH can create high-quality partitions of the dataset, our approach returns better candidate sets (i.e., Our candidate sets contain more of the k-Nearest Neighbors for any given query point.)  for query points since we use multiple complementary partitions per dataset through ensembling.

\subsubsection{Comparing with tree-based methods}
\label{tree-methods}

\begin{figure*}[h]
     \centering
     \begin{subfigure}[b]{0.45\textwidth}
         \centering
         \includegraphics[width=\textwidth]{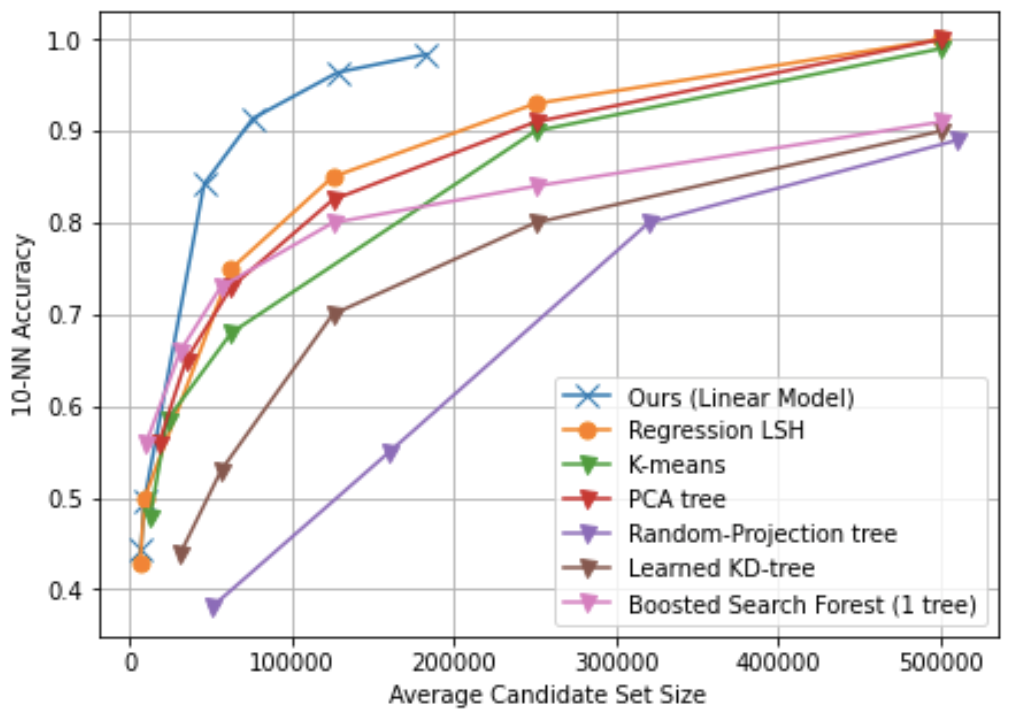}
         \caption{SIFT, 1024 bins}
         \label{sift-1024}
     \end{subfigure}
    \hspace{5mm}%
     \begin{subfigure}[b]{0.45\textwidth}
         \centering
         \includegraphics[width=\textwidth]{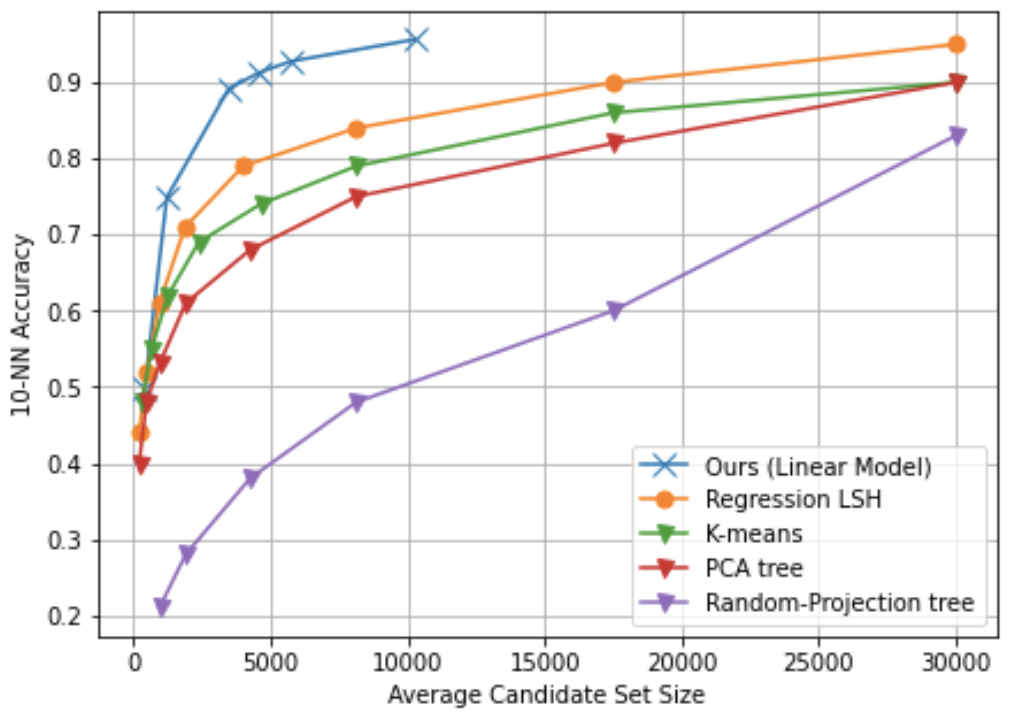}
         \caption{MNIST, 1024 bins}
         \label{mnist-1024}
     \end{subfigure}
  
     \captionsetup{justification=centering}
    \caption{Comparing our method with binary decision trees that use hyperplane partitions. X-axis: number of candidates retrieved in the candidate set. Y axis: 10-NN accuracy (Up and to the left is better).}
    \label{fig:tree_graphs}
\end{figure*}

We compare the performance of our approach with baselines that use hyperplanes to partition the dataset (Figure ~\ref{fig:tree_graphs}). In this setting, we use binary decision trees up to depth 10, which correspond to the dataset being divided recursively into $2^{10} = 1024$ bins for each of the methods compared. We note that our method, using a logistic regression learner, significantly outperforms Regression LSH without any ensembling. This is especially true in the high accuracy regime, where in SIFT, for instance, to obtain a 10-NN accuracy of about $98\%$, our approach returns candidate set sizes that are about \textbf{60\%} smaller than the best performing baselines.

\begin{table}[]
    \centering
    \begin{tabular}{c|c}
        Method & Decrease in candidate set size for 10-NN search \\
        \hline
        Neural LSH & 33\% \\
        K-means & 38\%\\
    \end{tabular}
    \captionsetup{justification=centering}
    \caption{Relative decrease in candidate size when searching for 10-Nearest Neighbors in SIFT, maintaining 10-NN accuracy of 85\% in Figure ~\ref{sift-16}}
    \label{table-compare}
\end{table}

\subsubsection{Comparing with non-learning ANNS methods}
\begin{figure*}[h]
     \centering
     \begin{subfigure}[b]{0.45\textwidth}
         \centering
         \includegraphics[width=\textwidth]{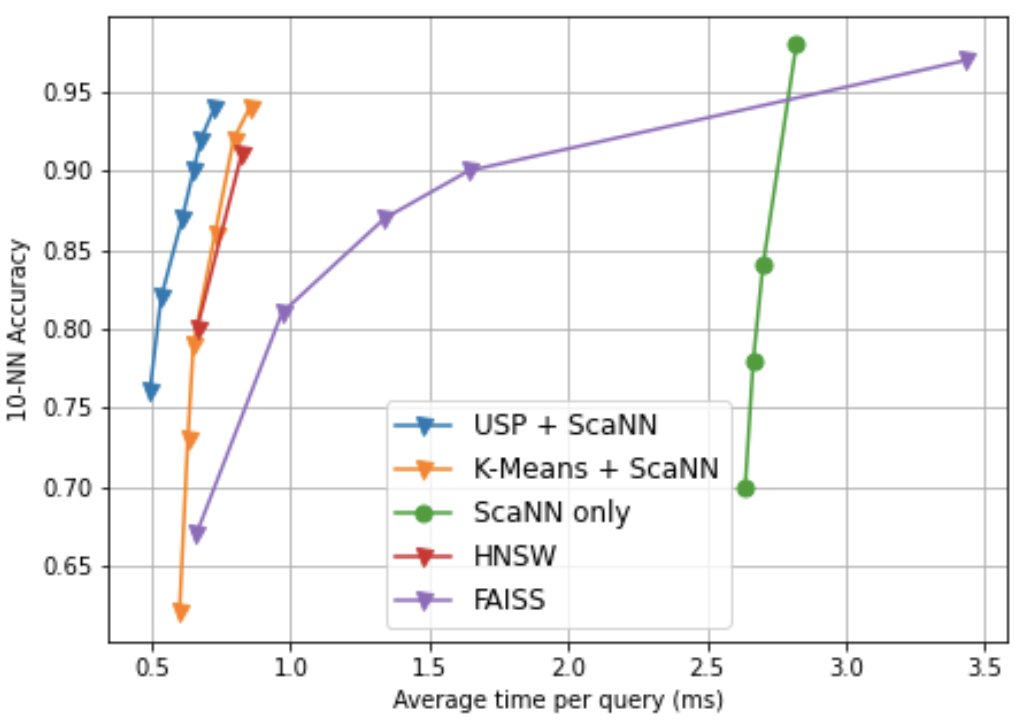}
         \caption{SIFT}
         \label{sift-scann}
     \end{subfigure}
    \hspace{5mm}%
     \begin{subfigure}[b]{0.45\textwidth}
         \centering
         \includegraphics[width=\textwidth]{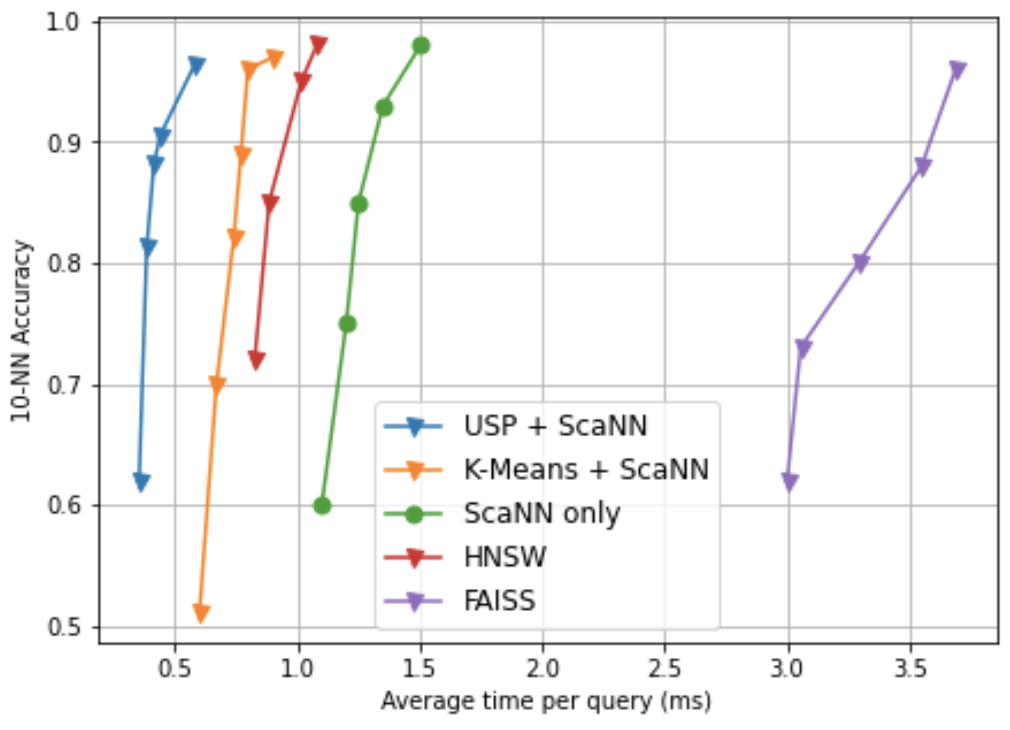}
         \caption{MNIST}
         \label{mnist-scann}
     \end{subfigure}
  
     \captionsetup{justification=centering}
    \caption{Using our partitioning method to enhance ScaNN's performance (Up and to the left is better). ScaNN + Ours outperforms commonly used previous best ANNS baselines.}
    \label{fig:scann_graphs}
\end{figure*}

\begin{table*}[t]
    \centering
    \begin{tabular}{cccc}
        \textbf{Our Approach} & \textbf{DBSCAN} & \textbf{K-means} & \textbf{Spectral clustering} \\
        \hline
        \begin{subfigure}[b]{0.20\textwidth}
         \centering
         \includegraphics[width=\textwidth]{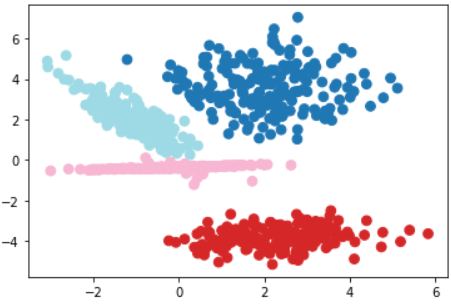}
        \end{subfigure} & \begin{subfigure}[b]{0.20\textwidth}
         \centering
         \includegraphics[width=\textwidth]{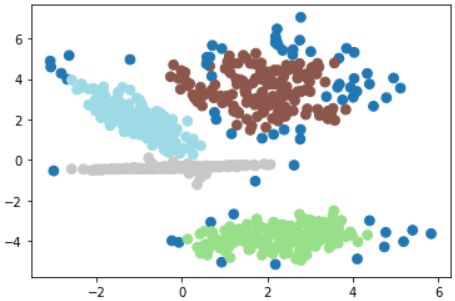}
        \end{subfigure} & \begin{subfigure}[b]{0.20\textwidth}
         \centering
         \includegraphics[width=\textwidth]{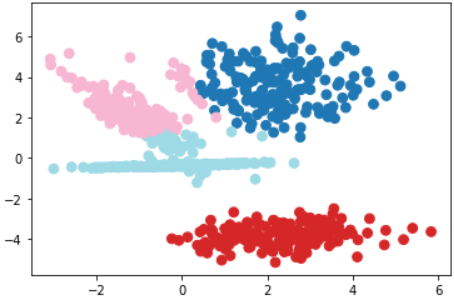}
        \end{subfigure} & \begin{subfigure}[b]{0.20\textwidth}
         \centering
         \includegraphics[width=\textwidth]{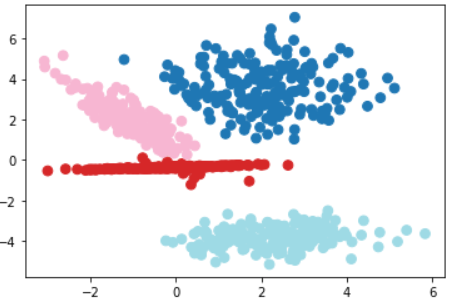}
        \end{subfigure} \\
     
      \begin{subfigure}[b]{0.20\textwidth}
         \centering
         \includegraphics[width=\textwidth]{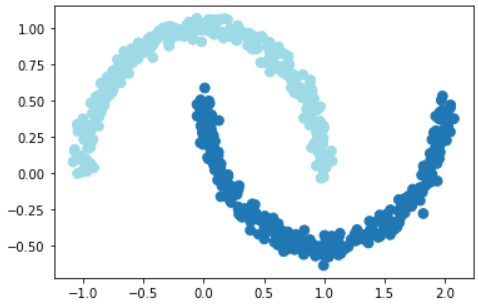}
     \end{subfigure} & \begin{subfigure}[b]{0.20\textwidth}
         \centering
         \includegraphics[width=\textwidth]{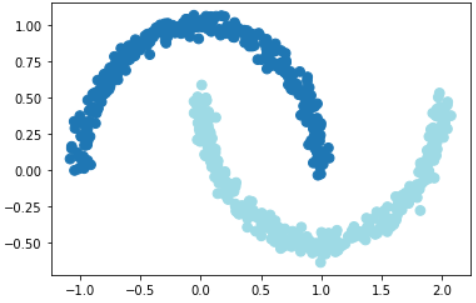}
     \end{subfigure} &  \begin{subfigure}[b]{0.20\textwidth}
         \centering
         \includegraphics[width=\textwidth]{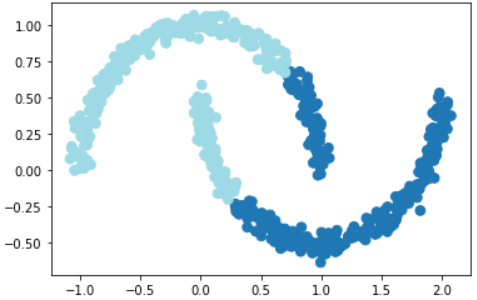}
     \end{subfigure} & \begin{subfigure}[b]{0.20\textwidth}
         \centering
         \includegraphics[width=\textwidth]{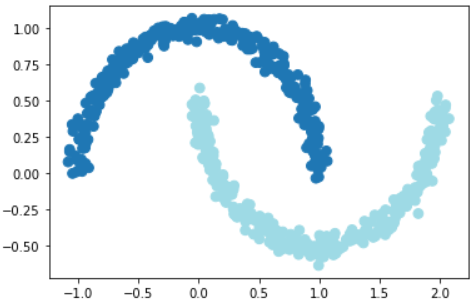}
     \end{subfigure} \\
     \begin{subfigure}[b]{0.20\textwidth}
         \centering
         \includegraphics[width=\textwidth]{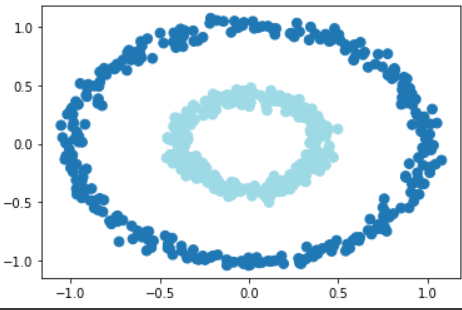}
     \end{subfigure} & \begin{subfigure}[b]{0.20\textwidth}
         \centering
         \includegraphics[width=\textwidth]{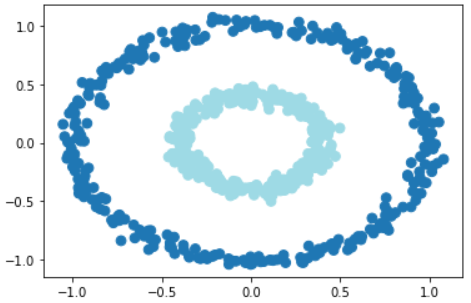}
     \end{subfigure} & \begin{subfigure}[b]{0.20\textwidth}
         \centering
         \includegraphics[width=\textwidth]{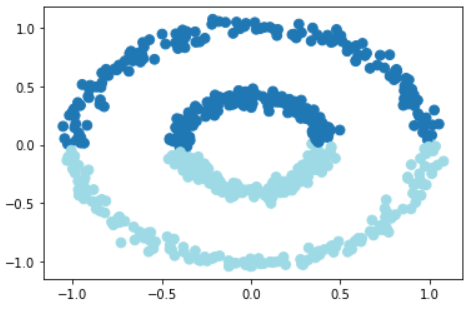}
     \end{subfigure} & \begin{subfigure}[b]{0.20\textwidth}
         \centering
         \includegraphics[width=\textwidth]{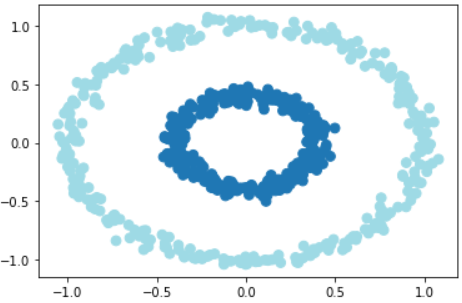}
     \end{subfigure} \\
     
    \end{tabular}
    \captionsetup{justification=centering}
    \caption{Comparing common clustering algorithms to our space-partitioning approach.}
    \label{table-clusters}
\end{table*}
 
\revise{In this set of experiments, we demonstrate the ubiquitous effectiveness of our partitioning approach in improving the performance of non-learning ANNS approaches.
We incorporated our partitioning approach in the best-performing ANNS method ScaNN. We first partition the data using our approach, where we split the dataset into a predetermined number of bins. Then, for a given query point $q$, we use our trained model to return a candidate set of points that are likely to be near $q$. Finally, we use ScaNN to search for the k-NNs of $q$ from its candidate set. In particular, we use ScaNN's novel anisotropic quantization method to speed up this search.
We term this pipeline as \emph{USP + ScaNN} algorithm, where USP refers to our proposed Unsupervised Space Partitioning approach. We show the effectiveness of this approach by comparing \emph{USP + ScaNN} with vanilla ScaNN (i.e., ScaNN without any data partitioning algorithm run beforehand), ScaNN with K-means tree partitioning (termed as \emph{K-means + ScaNN}, where K-means trees partition the dataset before running ScaNN),  HNSW, and FAISS. Figure ~\ref{fig:scann_graphs} outlines the results of our experiments. On average, the experiments show a $40\%$ speedup in 10-NN retrieval times compared to the best-performing approach, K-means + ScaNN.}

\subsection{Comparison with clustering methods}

\revise{The previous experiments show that our partitioning algorithm generates superior partitions compared to state-of-the-art partitioning baselines. Clustering algorithms (such as K-means clustering) split datasets into clusters and thus create partitions. We can similarly use our algorithm to create clusters of the dataset in an unsupervised manner. We show that the clusters created from our algorithm are better than the most commonly used clustering algorithms.}

We show the visualization of several 2D standard datasets (\emph{moon} and \emph{circles}) from scikit learn ~\cite{scikit-learn}, which are often used to determine the pitfalls of clustering algorithms. We also test with another sample dataset generated using \emph{make\_classification} from scikit learn with four clusters, which is challenging for many clustering algorithms. We compare our approach with common clustering algorithms DBSCAN ~\cite{ester_density-based_1996}, Spectral clustering ~\cite{ng_spectral_2001}, and K-means clustering in Table ~\ref{table-clusters}, where we show that our clustering performance is optimal for the test datasets. 
\revise{The results show that our approach successfully outputs the most natural clustering regardless of the shape of the data distribution.}

We note that even though spectral clustering achieves a similar quality clustering as ours, we cannot scale spectral clustering efficiently to large and high-dimensional datasets. Thus, our proposed partitioning approach can be a strong alternative to commonly used clustering techniques for high-dimensional datasets.

\section{Conclusions}\label{conclusion}

This paper proposes an end-to-end unsupervised learning framework that couples partitioning and learning to solve the ANNS problem in a single step. To facilitate the above, we propose a multi-objective custom loss function that guides the neural network (or any other learning model) to partition the space suitable for providing high-quality answers for ANNS. To further improve the performance, we propose an ensembling technique by adding varying input weights to the loss function to train multiple models and enhance search quality. Our experimental evaluation shows that our method beats the state-of-the-art learning-based ANNS approach while using fewer parameters and shorter offline training times on several benchmark datasets. \revise{We also show that our unsupervised partitioning approach boosts the current best-performing ANNS method, ScaNN, by 40\%.}
The code base of this paper is available at \url{https://github.com/abrar-fahim/Neural-Partitioner}.

\textbf{Acknowledgments:} This work is done at DataLab (datalab.buet.io), Dept of CSE, BUET. Muhammad Aamir Cheema is supported by ARC FT180100140.

\balance
\bibliographystyle{ACM-Reference-Format} 
\bibliography{8.references}

\end{document}